\documentclass[submission]{eptcs}
 
\usepackage{booktabs}
\usepackage{paralist}
\usepackage{amsmath}
\usepackage{xspace}

\providecommand{\event}{FVAV'17}

\newcommand{\relu}{\text{ReLU}\xspace{}}

\newtheorem{definition}{Definition}

\usepackage{tikz,ifthen,pgfplots}
\usetikzlibrary{arrows,trees,backgrounds,automata,shapes,decorations,plotmarks,fit,calc,positioning,shadows,chains}
\tikzstyle{every pin edge}=[<-,shorten <=1pt]
\tikzstyle{neuron}=[circle,fill=black!25,minimum size=17pt,inner sep=0pt]
\tikzstyle{input neuron}=[neuron, fill=green!50]
\tikzstyle{output neuron}=[neuron, fill=red!50]
\tikzstyle{hidden neuron}=[neuron, fill=blue!50]
\tikzstyle{annot} = [text width=4em, text centered]
\usetikzlibrary{calc}

\title{Towards Proving the Adversarial Robustness of
\\Deep Neural Networks}

\author{
Guy Katz, Clark Barrett, David L. Dill, Kyle Julian and Mykel J. Kochenderfer
\institute{Stanford University}
\email{\{guyk, clarkbarrett, dill, kjulian3, mykel\}@stanford.edu}
}

\def\titlerunning{Towards Proving Adversarial Robustness of Deep Neural Networks}
\def\authorrunning{G. Katz et al.}

\begin{document}
\maketitle

\begin{abstract}
Autonomous vehicles are highly complex systems, required to function
reliably in a wide variety of situations.
Manually crafting software controllers for these vehicles is
difficult, but there has been some success in using \emph{deep neural
  networks} generated using machine-learning. However, deep neural
networks are opaque to human 
engineers, rendering their correctness very difficult to prove
manually; and existing automated techniques, which were not designed
to operate on neural networks, fail to scale to large systems.
This paper focuses on proving the adversarial robustness of deep neural
networks, i.e. proving 
that small perturbations to a correctly-classified input to the
network cannot cause it to be misclassified.
We describe some of our recent and ongoing work on verifying the adversarial
robustness of networks,
and discuss some of the open questions we have encountered and how
they might be addressed. 
\end{abstract}

\section{Introduction}
Designing software controllers for autonomous vehicles is
a difficult and error-prone task. A main
cause of this difficulty is that, when deployed, 
autonomous vehicles may encounter a wide variety of situations 
and are required to perform reliably in each of them.
The enormous space of possible situations makes it 
nearly impossible for a human engineer to anticipate every
corner-case.

Recently, \emph{deep neural networks} (\emph{DNNs}) have
emerged as a way to effectively create complex software. 
Like other machine-learning generated
systems, DNNs are created by observing a finite set of input/output
examples of the correct behavior of
the system in question,
and extrapolating from them a software artifact capable of handling previously unseen situations. 
DNNs have proven remarkably useful in many applications,
 including  
 including speech
recognition~\cite{HiDeYuDaMoJaSeVaNgSaKi12}, image
classification~\cite{KrSuHi12}, and game
playing~\cite{SiHuMaGuSiVaScAnPaLaDi16}. There has also been a surge
of interest in using them as controllers in autonomous 
vehicles such as automobiles~\cite{BoDeDwFiFlGoJaMoMuZhZhZhZi16}
and aircraft~\cite{JuLoBrOwKo16}.

The intended use of DNNs in autonomous vehicles raises many questions
regarding the certification of such systems.  Many of the common
practices aimed at increasing software reliability --- such as code
reviews, refactoring, modular designs and manual proofs of correctness ---
simply cannot be applied to DNN-based software. Further,
existing automated verification tools are typically ill-suited to
reason about DNNs, and they fail to scale to anything larger than toy
examples~\cite{PuTa10,PuTa12}. Other approaches use various forms of
approximation~\cite{BaIoLaVyNoCr16,HuKwWaWu16} to
achieve scalability, but using approximations may not meet the
certification bar for
safety-critical systems. Thus, it is clear that new methodologies and
tools for scalable verification of DNNs are sorely needed.  

We focus here on a specific kind of desirable property of DNNs, called
\emph{adversarial robustness}. Adversarial robustness measures a
network's resilience against \emph{adversarial
  inputs}~\cite{SzZaSuBrErGoFe13}: inputs that 
are produced by taking inputs that are correctly classified by the DNN
and perturbing them slightly, in a way that causes them to be
misclassified by the network.
 For example, for a DNN for image recognition such
examples can correspond to slight distortions in the input image that
are invisible to the human eye, but cause the network to assign
the image a completely different classification.
 It has been observed that many
state-of-the-art DNNs are highly vulnerable to adversarial inputs, 
and several highly effective techniques have been devised for finding
such inputs~\cite{CaWa17,GoShSz14}. Adversarial attacks
can be carried out in the real world~\cite{KuGoBe16}, and  
 thus constitute a source of concern for autonomous vehicles using
 DNNs --- making it 
 desirable to verify that these DNNs are robust.

In a recent paper~\cite{KaBaDiJuKo17}, we proposed a new decision procedure, called
Reluplex, designed to solve systems of linear equations with certain
additional, non-linear constraints. In particular, neural networks and
various interesting properties thereof can be encoded as input to
Reluplex, and the properties can then be proved (or disproved, in
which case a counter
example is provided). We used Reluplex to verify various properties of 
a prototype DNN implementation of the next-generation Airborne
Collision Avoidance Systems (ACAS Xu), which is currently being developed by
the Federal Aviation Administration (FAA)~\cite{JuLoBrOwKo16}.   

This paper presents some of our ongoing efforts along this line
of work, focusing on adversarial robustness properties. We study
different kinds of robustness properties and practical
considerations for proving them on real-world networks. We also present
some initial results on proving these properties for the ACAS Xu
networks. Finally, we discuss some of the open questions we have
encountered and our plans for addressing them in the future.

The rest of this paper is organized as follows. We briefly provide some needed
background on DNNs and on Reluplex in Section~\ref{sec:background},
followed by a discussion of adversarial robustness in
Section~\ref{sec:adversarialRobustness}. We continue with a discussion of
our ongoing research
and present some initial experimental results
in Section~\ref{sec:movingForward}, and
conclude with Section~\ref{sec:conclusion}.

\section{Background}
\label{sec:background}
  
\subsection{Deep Neural Networks}
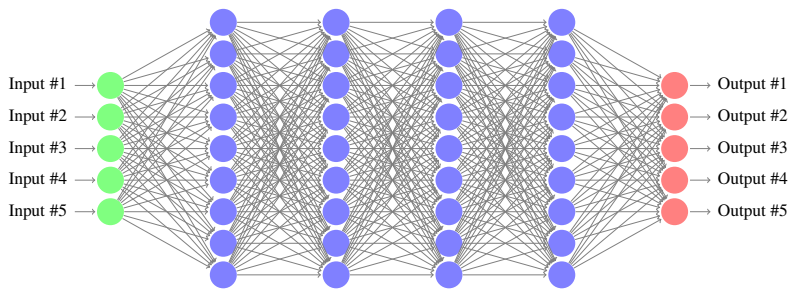
\begin{figure}[htp]
\begin{center}
\scalebox{0.6}{
\def\layersep{2.5cm}
\def\vertSepFactory{0.7}
\begin{tikzpicture}[shorten >=1pt,->,draw=black!50, node distance=\layersep]
    \foreach \name / \y in {1,...,5}
        \node[input neuron, pin=left:Input \#\y] (I-\name) at (0,-\vertSepFactory * \y) {};

    \foreach \name / \y in {1,...,9}
        \path[yshift=1.4cm]
            node[hidden neuron] (H1-\name) at (1*\layersep,-\vertSepFactory * \y cm) {};

    \foreach \name / \y in {1,...,9}
        \path[yshift=1.4cm]
            node[hidden neuron] (H2-\name) at (2*\layersep,-\vertSepFactory * \y cm) {};

    \foreach \name / \y in {1,...,9}
        \path[yshift=1.4cm]
            node[hidden neuron] (H3-\name) at (3*\layersep,-\vertSepFactory * \y cm) {};

    \foreach \name / \y in {1,...,9}
        \path[yshift=1.4cm]
            node[hidden neuron] (H4-\name) at (4*\layersep,-\vertSepFactory * \y cm) {};

    \foreach \name / \y in {1,...,5}
        \node[output neuron,pin={[pin edge={->}]right:Output \#\y}]
        (O-\name) at (5*\layersep, -\vertSepFactory * \y cm) {};

    \foreach \source in {1,...,5}
        \foreach \dest in {1,...,9}
            \path (I-\source) edge (H1-\dest);

    \foreach \source in {1,...,9}
        \foreach \dest in {1,...,9}
            \path (H1-\source) edge (H2-\dest);

    \foreach \source in {1,...,9}
        \foreach \dest in {1,...,9}
            \path (H2-\source) edge (H3-\dest);

    \foreach \source in {1,...,9}
        \foreach \dest in {1,...,9}
            \path (H3-\source) edge (H4-\dest);

    \foreach \source in {1,...,9}
        \foreach \dest in {1,...,5}
            \path (H4-\source) edge (O-\dest);

\end{tikzpicture}
}
\caption{A DNN with 5 input nodes (in green), 5 output
  nodes (in red), and 36 hidden nodes (in blue). The network has 6 layers.
}
\label{fig:fullyConnectedNetwork}
\end{center}
\end{figure}

Deep neural networks (DNNs) consist of a set of nodes (``neurons''),
organized in a layered structure. Nodes in 
the first layer are called input nodes, nodes in the last layer are
called output nodes, and nodes in the intermediate layers are called
hidden nodes. 
An example appears in
Fig.~\ref{fig:fullyConnectedNetwork} (borrowed from~\cite{KaBaDiJuKo17}).

Nodes are connected to nodes from the preceding layer by weighted
edges, and are each assigned a bias value. An evaluation of the DNN is
performed as follows. First, the 
input nodes are assigned values (these can correspond, e.g., to user
inputs or sensor readings). Then, the network is evaluated
layer-by-layer: in each layer the values of the nodes are calculated
by (i) computing a weighted sum of values from the previous layer,
according to the weighted edges;
(ii) adding each node's bias value to the weighted sum; and (iii)
applying a predetermined activation function to the result of (ii). The value returned by the
activation function becomes the value of the node, and this process is
propagated layer-by-layer until the network's output values are computed. 

This work focuses on DNNs using a particular
kind of activation function, called a \emph{rectified linear unit}
(\emph{ReLU}). 
The ReLU function is given by the piecewise linear formula 
$\relu{}(x)=\max(0,x)$, i.e., positive values are unchanged and negative
values are changed to $0$. When applied to a positive value, we say that
the ReLU is in the \emph{active} state; and when applied to a non-positive
value, we say that it is in the \emph{inactive} state.
ReLUs are very widely used in practice~\cite{KrSuHi12,MaHaNg13}, and it has been suggested that the
 piecewise linearity that they introduce allows DNNs to 
generalize well to new inputs~\cite{GlBoBe11, FoBeCu16, JaKaLe09, NaHi10}.
  
A DNN $N$ is referred to as a \emph{classifier} if it is associated with a
set of labels $L$, such that each output node of $N$ corresponds to a
specific output label. For a given input $\vec{x}$ and label $\ell\in
L$, we refer to the value of $\ell$'s output node as the
\emph{confidence} of $N$ that $\vec{x}$ is labeled $\ell$, and denote
this value by $C(N,\vec{x},\ell)$.
 An input $\vec{x}$ is said to be classified to label
$\ell\in L$, denoted $N(\vec{x})=\ell$, if $C(N,\vec{x},\ell)> C(N,\vec{x},\ell')$ for
all $\ell'\neq \ell$.

\subsection{Verifying Properties of Neural Networks}

A DNN can be regarded as a collection of linear equations, with the
additional ReLU constraints. Existing
verification tools capable of handling these kinds of constraints
include linear programming (LP) solvers and satisfiability modulo
theories (SMT) solvers, and indeed past research has focused on using
these tools~\cite{BaIoLaVyNoCr16,HuKwWaWu16,PuTa10,PuTa12}.
As for the properties being verified, we restrict our attention to
properties that can be expressed as linear constraints over the DNN's input and
output nodes. Many properties of interest seem to fall into this category,
including adversarial robustness~\cite{KaBaDiJuKo17}.

Unfortunately, this verification problem is
NP-complete~\cite{KaBaDiJuKo17}, making it theoretically difficult. It
is also difficult in practice, with modern solvers scaling only to
very small examples~\cite{PuTa10,PuTa12}. 
Because problems involving 
only linear constraints are fairly easy to solve, many solvers 
handle the ReLU constraints by transforming the input query into a sequence of pure linear
sub-problems, such that the original query is satisfiable if and only if
at least one of the sub-problems is satisfiable. This transformation
is performed by \emph{case-splitting}:
given a query involving         
$n$ ReLU constraints, the linear sub-problems are obtained by fixing
each of the ReLU constraints in either the active or inactive state
(recall that ReLU constraints are piecewise linear). 
Unfortunately, this entails exploring every possible
combination of active/inactive ReLU states, 
meaning that the solver needs to check
$2^n$ linear sub-problems in the worst case. This quickly becomes a crucial
bottleneck when $n$ increases.

In a recent paper, we proposed a new algorithm, called Reluplex,
capable of verifying DNNs that are an order-of-magnitude larger than
was previously possible~\cite{KaBaDiJuKo17}. The key insight that led to this improved
scalability was a lazy treatment of the ReLU constraints: instead of
exploring all possible combinations of ReLU activity or inactivity,
Reluplex temporarily ignores the ReLU constraints and attempts to
solve just the linear portion of the
problem. Then, by deriving variable bounds from the linear equations
that it explores, Reluplex is often able to deduce that some of the
ReLU constraints are \emph{fixed} in either the active or inactive
case, which greatly reduces the amount of case-splitting that it later
needs to perform. This has allowed us to use Reluplex to verify
various properties of the DNN-based implementation of the ACAS Xu
system: a family of 45 DNNs, each with 300 ReLU nodes.

\section{Adversarial Robustness}
\label{sec:adversarialRobustness}

A key challenge in software verification, and in particular in DNN
verification, is obtaining a specification against which the software
can be verified. One solution is to manually develop such
properties on a per-system basis, but we
can also focus on properties that are desirable for every
network. Adversarial robustness properties fall into this category:
they express the requirement that the network behave smoothly,
i.e. that small input perturbations should not cause major spikes in
the network's
output. Because DNNs are trained over a finite set of inputs/outputs,
this captures our desire to ensure that the network behaves ``well''
on inputs that were neither tested nor trained on. If adversarial
robustness is determined to be too low in certain parts of the input
space, the DNN may be retrained to increase its robustness~\cite{GoShSz14}.

We begin with a common definition for \emph{local adversarial
  robustness}~\cite{BaIoLaVyNoCr16,HuKwWaWu16,KaBaDiJuKo17}:  
\begin{definition} 
\label{def:localRobustness}A DNN $N$ is $\delta$-locally-robust at point
$\vec{x_0}$ iff
\[
\forall \vec{x}.\quad \| \vec{x}-\vec{x_0}\| \leq \delta \quad\Rightarrow\quad N(\vec{x})=N(\vec{x_0})
\]
\end{definition}
Intuitively, Definition~\ref{def:localRobustness} states that for
input $\vec{x}$ that is very close to $\vec{x_0}$, the network assigns to $\vec{x}$ the
same label that it assigns to $\vec{x_0}$; ``local'' thus refers to a local
neighborhood around $\vec{x_0}$. Larger values of $\delta$ imply larger
neighborhoods, and hence better robustness.
Consider, for instance, a DNN for
image recognition: $\delta$-local-robustness can then capture the fact
that slight perturbations of the input image, i.e. perturbations so
small that a human observer would fail to detect them, should not
result in a change of label.

There appear to be two main drawbacks to using
Definition~\ref{def:localRobustness}:
\begin{inparaenum}[(i)]
\item 
The property is checked for individual input points in an infinite
input space, and it does not necessarily carry over to other points
that are not checked. This issue may be partially mitigated by
testing points drawn from some random distribution thought to
represent the input space.
\item
 For each point $\vec{x_0}$ we need to specify the minimal
acceptable value of $\delta$. Clearly, these values can vary between
different input points: for example, a point 
 deep within a region
that is expected to be labeled $\ell_1$ should have high
robustness, whereas for a point closer to the boundary between two
labels $\ell_1$ and $\ell_2$ even a tiny $\delta$ may be
acceptable. We note that given a point $\vec{x_0}$ and a solver such as Reluplex, one
can perform a binary search and find the largest $\delta$ for which $N$ is
$\delta$-locally-robust at $\vec{x_0}$ (up to a desired precision).
\end{inparaenum}

In order to overcome the need to specify each individual $\delta$
separately,  in~\cite{KaBaDiJuKo17} we
proposed an alternative approach, using the notion of \emph{global
  robustness}:

\begin{definition}
\label{def:globalRobustness}
A DNN $N$ is $(\delta,\epsilon)$-globally-robust in input region $D$
iff
\[
\forall \vec{x_1},\vec{x_2}\in D.\quad \| \vec{x_1}-\vec{x_2}\| \leq \delta \quad\Rightarrow\quad
\forall \ell\in L.\quad |C(N,\vec{x_1},\ell)-C(N,\vec{x_2},\ell)|< \epsilon
\]
\end{definition}
Definition~\ref{def:globalRobustness} addresses the two shortcomings of
Definition~\ref{def:localRobustness}. First, 
it considers an input 
domain $D$ instead of a specific point $\vec{x_0}$, allowing it to cover
infinitely many points (or even the entire input space) in a single
query, with $\delta$ and $\epsilon$ defined once for the entire domain. Also, it is better 
suited for handling input points that lay on the boundary between two labels:
this definition now only requires that two $\delta$-adjacent points
are classified in a similar (instead of identical) way, in the sense
that there are no spikes greater than $\epsilon$ in the levels of confidence
that the network assigns to each label for these
points. Here it is desirable to have a large $\delta$ (for large
neighborhoods) and a small $\epsilon$ (for small spikes), although it
is expected that the two parameters will be mutually dependent.

Unfortunately, global robustness appears to be
significantly harder to check, as we discuss next.

\subsection{Verifying Robustness using Reluplex}
\label{sec:verifyingWithReluplex}

Provided that the distance metrics in use can be
expressed as a combination of linear constraints and ReLU operators
($L_1$ and $L_\infty$ fall into this category),
$\delta$-local-robustness and $(\delta,\epsilon)$-global-robustness
properties can be encoded as Reluplex inputs.
For the local robustness case, 
the input constraint $\| \vec{x}-\vec{x_0}\|\leq \delta$ is encoded
directly as a set of linear equations and variable bounds, and the robustness property is negated
and encoded as 
\[
\bigvee_{\ell\neq N(\vec{x_0})}N(\vec{x})=\ell
\]
Thus, if Reluplex finds a variable assignment that satisfies the
query, this assignment
constitutes a counter-example $\vec{x}$ that
violates the property, i.e., $\vec{x}$ is $\delta$-close to $\vec{x_0}$ but has a
label different from that of $\vec{x_0}$. If Reluplex discovers that the
query is unsatisfiable, then the network is guaranteed to be $\delta$-local-robust at
$\vec{x_0}$. 

Encoding $(\delta,\epsilon)$-global-robustness is more difficult
because neither $\vec{x_1}$ nor $\vec{x_2}$ is fixed. It is performed
by encoding 
two copies of the network, denoted $N_1$ and $N_2$, such that $\vec{x_1}$ is
the input to $N_1$ and $\vec{x_2}$ is the input to $N_2$. We again encode the
constraint $\| \vec{x_1} - \vec{x_2} \|\leq \delta$ as a set of linear equations, and the robustness property
is negated and encoded as
\[
\bigvee_{\ell\in L}| C(N_1,\vec{x_1},\ell) - C(N_2,\vec{x_2},\ell)  |\geq\epsilon
\]
As before, if the query is unsatisfiable then the property holds,
whereas a satisfying assignment constitutes a counter-example.

While both kinds of queries can be encoded in Reluplex, global
robustness is significantly harder to prove than its local
counterpart. The main reason is the technique mentioned in
Section~\ref{sec:verifyingWithReluplex}, 
which allows Reluplex
to achieve scalability by determining that
certain ReLU constraints are fixed at either the active or inactive
state.
When checking local robustness, the
network's input nodes are restricted to a small neighborhood around
$\vec{x_0}$, and this allows Reluplex to discover that many ReLU
constraints are fixed;
 whereas
the larger domain $D$ used for global robustness queries tends to
allow fewer ReLUs to be eliminated, which entails additional
case-splitting and slows Reluplex down.
Also, as previously explained, encoding a
global-robustness property entails encoding two identical copies of
the DNN in question. This doubles the number of variables and ReLUs
that Reluplex needs to handle, leading to slower
performance. Consequently, our implementation of Reluplex can
currently verify the local adversarial robustness of DNNs with several
hundred nodes, whereas global robustness is limited to DNNs with a few dozen
nodes.

\section{Moving Forward}
\label{sec:movingForward}
  
A significant question in moving forward is on which definition of adversarial
robustness to focus. The advantages of using
$(\delta,\epsilon)$-global-robustness are clear, but the present
state-of-the-art seems insufficient for verifying it;
whereas $\delta$-local-robust is more feasible but requires a high
degree of manual fine tuning.
We suggest to focus for now on the following hybrid definition, which
is an enhanced
version of local robustness:
\begin{definition}
\label{def:hybridLocalRobustness}
A DNN $N$ is $(\delta,\epsilon)$-locally-robust at point
$\vec{x_0}$ iff
\[
\forall \vec{x}.\quad \| \vec{x}-\vec{x_0}\| \leq \delta \quad\Rightarrow\quad
\forall \ell\in L.\quad |C(N,\vec{x},\ell) - C(N,\vec{x_0},\ell)|<\epsilon
\]
\end{definition}
The encoding of $(\delta,\epsilon)$-local-robustness properties as
inputs to Reluplex is similar to the previous cases:
the constraint $\| \vec{x} - \vec{x_0} \|\leq \delta$ is encoded as a
set of linear equations and variable bounds, and the robustness property
is negated and encoded as
\[
\bigvee_{\ell\in L}| C(N,\vec{x},\ell) - C(N,\vec{x_0},\ell)  |\geq\epsilon
\]
Definition~\ref{def:hybridLocalRobustness} is still local in nature,
which means that testing it using Reluplex does not require encoding two copies
of the network. It also allows ReLU elimination, which affords some scalability 
(see Table~\ref{table:evaluation} for some initial results). Finally,
this definition's notion of robustness is based on the difference in confidence levels, as
opposed to a different labeling, making it more easily applicable to any
input point, even if it is close to a boundary between two labels. Thus, we
believe it is superior to Definition~\ref{def:localRobustness}.
An open problem is how to determine the finite set of points to be
tested, and the $\delta$ and $\epsilon$ values to test. (Note that it
may be possible to use the same $\delta$ and $\epsilon$ values for all
points tested, reducing the amount of manual work required.)

Another important challenge in moving forward is
scalability. Currently, Reluplex is able to handle DNNs with several
hundred nodes, but many real-world DNNs are much larger than that. Apart
from improving the Reluplex heuristics and implementation, we believe that parallelization will
play a key role here. Verification of robustness properties, both local and
global, naturally lends itself to parallelization. In the local case,
testing the robustness of $n$ input points  can be performed
simultaneously using $n$ machines; and even in the global case,
an input domain $D$ can be partitioned into $n$ sub domains $D_1,\ldots,D_n$, each of
which can be tested separately. The experiment described in
Table~\ref{table:evaluation} demonstrates the benefits of
parallelizing $(\delta,\epsilon)$-local-robustness testing even further: apart from testing each point on a separate
machine, for each point the disjuncts in
the encoding of Definition~\ref{def:hybridLocalRobustness} can also be checked in parallel. The
improvement in performance is evident, emphasizing the potential
benefits of pursuing this direction further.

 We believe parallelization can be made even more efficient in this context by
means of two complementary directions:
\begin{enumerate}
\item \textbf{Prioritization.}
When testing the (local or global) robustness of a DNN, we
can stop immediately once a violation has been found. Thus,
prioritizing the points or input domains and starting from those in
which a violation is most likely to occur could serve to reduce
execution time. Such prioritization could be made possible by
numerically analyzing the network prior to verification, identifying
input regions
in which there are steeper fluctuations in the output values, and focusing
on these regions first.

\item \textbf{Information sharing across nodes.} 
As previously mentioned, a key aspect of the scalability of Reluplex
is its ability to determine that certain ReLU constraints are
fixed in either the active or inactive case. When running multiple
experiments, these conclusions could potentially be shared between
executions, improving performance. Of course, great care will need to be
taken, as a ReLU that is fixed in one input domain may not be fixed (or
may even be fixed in the other state) in another domain. 
\end{enumerate} 

Finally, we believe it would be important to come up with automatic
techniques for choosing the input points (in the local case) or
domains (in the global case) to be tested, and the corresponding
$\delta$ and $\epsilon$ parameters. These techniques would likely take
into account the distribution of the inputs in the network's training set. In the global
case, domain selection could be performed in a way that would 
optimize the verification process, by selecting domains in
which ReLU constraints are fixed in the active or inactive state.

\begin{table}
\centering
\caption{
Checking the $(\delta,\epsilon)$-local-robustness of one of the ACAS
Xu DNNs~\cite{KaBaDiJuKo17} at 5 arbitrary input points, for different values of $\epsilon$
(we fixed $\delta=0.018$ for all experiments). The \emph{Seq.} columns
indicate execution time (in seconds) for a sequential execution, and the
\emph{Par.} columns indicate execution time (in seconds) for a
parallelized execution using $5$ machines.
}
\scalebox{0.8}{
\begin{tabular}[htp]{c|clrrclrrclrr}
  \toprule
  Point 
  &&
    \multicolumn{3}{c}{$\epsilon = 0.01$}
  &&
     \multicolumn{3}{c}{$\epsilon = 0.02$}
  && 
     \multicolumn{3}{c}{$\epsilon = 0.03$}
  \\
  \cline{3-5} \cline{7-9} \cline{11-13}
  &&
     Robust? & Par. & Seq.
  &&
     Robust? & Par. & Seq.
  &&
     Robust? & Par. & Seq.
  \\
  \midrule
  1     && No  & 5   & 5    && No  & 785 & 7548 && Yes & 9145 & 38161  \\
  2     && Yes & 277 & 1272 && Yes & 248 & 989  && Yes & 191  & 747 \\
  3     && Yes & 103 & 460  && Yes & 134 & 480  && Yes & 93   & 400 \\
  4     && No  & 17  & 17   && Yes & 249 & 774  && Yes & 132  & 512 \\
  5     && Yes & 333 & 1479 && Yes & 259 & 1115 && Yes & 230  & 934 \\
  \bottomrule
\end{tabular}%
\label{table:evaluation}
}
\end{table}%

\section{Conclusion}
\label{sec:conclusion}

The planned inclusion of DNNs within autonomous vehicle controllers
poses a significant challenge for their certification. In particular,
it is becoming increasingly important to show that these DNNs are robust to
adversarial inputs. This challenge
can be addressed through verification, but the scalability of
state-of-the-art techniques is a limiting factor and dedicated
techniques and methodologies need to be developed for this
purpose. 

In~\cite{KaBaDiJuKo17} we presented the Reluplex
algorithm which is capable of proving DNN robustness in some cases. Still,
additional work is required to improve scalability.
We believe that by carefully phrasing the properties being
proved, and by intelligently applying parallelization, a significant
improvement can be achieved.

As a long-term goal, we speculate that this line of work could assist
researchers in verifying the dynamics of autonomous vehicle systems
that include a DNN-based controller.
In particular, it may be possible to
first formally prove that a DNN-based controller satisfies certain
properties, and then use these properties in analyzing the dynamics of
the system as a whole. Specifically, we plan to explore the
integration of Reluplex with reachability analysis techniques, for
both the offline~\cite{JaGhKoGaScZaPl15} and online~\cite{AlDo14} variants.

\medskip
\noindent
\textbf{Acknowledgements.} We thank 
Neal Suchy from the FAA and Lindsey Kuper from Intel 
 for their valuable comments and support.
This work was partially supported by 
 grants from the FAA and Intel.

\bibliographystyle{eptcs}

\end{document}